# PROBABILISTIC TAGGING WITH FEATURE STRUCTURES

André Kempe

University of Stuttgart, Institute for Computational Linguistics,
Azenbergstraße 12, 70174 Stuttgart, Germany,    kempe@ims.uni-stuttgart.de

**Abstract**

The described tagger is based on a hidden Markov model and uses tags composed of features such as part-of-speech, gender, etc. The contextual probability of a tag (state transition probability) is deduced from the contextual probabilities of its feature-value-pairs.

This approach is advantageous when the available training corpus is small and the tag set large, which can be the case with morphologically rich languages.

## 1 INTRODUCTION

The present article describes a probabilistic tagger based on a *hidden Markov model* (HMM) (Rabiner, 1990) and employs tags which are feature structures. Their features concern part-of-speech (POS), gender, number, etc. and have only atomic values.

Usually, the contextual probability of a tag (state transition probability) is estimated dividing a trigram frequency by a bigram frequency (second order HMM). With a large tag set resulting from the fact that the tags contain besides of the POS a lot of morphological information, and with only a small training corpus available, most of these frequencies are too low for an exact estimation of contextual probabilities.

Our feature structure tagger estimates these probabilities by connecting contextual probabilities of the single *feature-value-pairs* (fv-pairs) of the tags (cf. sec. 2).

Starting point for the implementation of the feature structure tagger was a second-order-HMM tagger (trigrams) based on a modified version of the Viterbi algorithm (Viterbi, 1967; Church, 1988) which we had earlier implemented in C (Kempe ,1994). There we modified the calculus of the contextual probabilities of the tags in the above-described way (cf sec. 4).

A test of both taggers under the same conditions on a French corpus[1] has shown that the feature structure tagger is clearly better when the available training corpus is small and the tag set is large but the tags are decomposable into relatively few fv-pairs. The latter can be the case with morphologically rich languages when the tags contain a lot of morphological information (cf. sec. 5).

## 2 MATHEMATICAL BACKGROUND

In order to assign tags to a word sequence, a HMM can be used where the tagger selects among all possible tag sequences the most probable one (Garside, Leech and Sampson, 1987; Church, 1988; Brown et al., 1989; Rabiner, 1990). The joint probability of a tag sequence $\bar{t} = t_0...t_{N-1}$ given a word sequence $\bar{w} = w_0...w_{N-1}$ is in the case of a second order HMM:

$$p(\bar{t}, \bar{w}) = \pi_{t_0 t_1} \cdot p(w_0|t_0) \cdot p(w_1|t_1) \cdot \prod_{i=2}^{N-1} \Big( p(w_i|t_i) \cdot p(t_i|\ t_{i-2}\ t_{i-1}) \Big) \quad (1)$$

The term $\pi_{t_0 t_1}$ stands for the initial state probability, i.e. the probability that the sequence begins with the first two tags. $N$ is the number of words in the sequence, i.e. the corpus size. The term $p(w_i|t_i)$ is the probability of a word $w_i$ in the context of the assigned tag $t_i$. It is called observation symbol probability (lexical probability) and can be estimated by:

$$p(w_i|t_i) = \frac{f(w_i\ t_i)}{f(t_i)} \quad (2)$$

The second order state transition probability (contextual probability) $p(t_i|\ t_{i-2}\ t_{i-1})$ in formula (1) expresses how probable it is that the tag $t_i$ appears in the context of its two preceding tags $t_{i-2}$ and $t_{i-1}$. It is usually estimated as the ratio of the frequency of the trigram $\langle t_{i-2}, t_{i-1}, t_i \rangle$ in a given training corpus to the frequency of the bigram $\langle t_{i-2}, t_{i-1} \rangle$ in the same corpus:

$$p(t_i|\ t_{i-2}\ t_{i-1}) = \frac{f(t_{i-2}\ t_{i-1}\ t_i)}{f(t_{i-2}\ t_{i-1})} \quad (3)$$

With a large tag set and a relatively small hand-tagged training corpus formula (3) has an important disadvantage: The majority of transition probabilities cannot be estimated exactly because most of the possible trigrams (sequences of three consecutive tags) will not appear at all or only a few times[2].

In our example we have a French training corpus of 10,000 words tagged with a set of 386 different tags which could form $386^3 = 57,512,456$ different trigrams, but because of the corpus size no more than 10,000-2 trigrams can appear. Actually, their number was only 4,815, i.e. 0.008 % of all possible

---

[1] I am much obliged to Achim Stein and Leo Wanner, Romance Dept., Univ. Stuttgart, Germany, for providing the corpus and a dictionary.

[2] A detailed description of problems caused by small and zero frequencies was given by Gale and Church (1989)

ones, because some of them appeared more than once (table 1).

| frequency range | number and percentage of trigrams in the range | |
|---|---|---|
| $\geq 128$ | 1 | (0.021 %) |
| 64 - 127 | 2 | (0.042 %) |
| 32 - 63 | 13 | (0.26 %) |
| 16 - 31 | 43 | (0.89 %) |
| 8 - 15 | 119 | (2.5 %) |
| 4-7 | 282 | (5.9 %) |
| 2-3 | 860 | (18 %) |
| 1 | 3,495 | (73 %) |
| sum | 4,815 | (100 %) |

Table 1: Trigram count from a French training corpus of 10,000 words

When we divide e.g. a trigram frequency 1 by a bigram frequency 2 according to formula (3) we get the probability $p=0.5$ but we cannot trust it to be exact because the frequencies it is based on are too small.

We can take advantage of the fact that the 386 tags are constituted by only 57 different fv-pairs concerning POS, gender, number, etc. If we consider probabilistic relations between single fv-pairs then we get higher frequencies (fig. 2) and the resulting probabilities are more exact.

From the equations

$$\{t_i\} = \{e_{i0} \cap e_{i1} \ldots \cap e_{i,n-1}\} = \left\{\bigcap_{k=0}^{n-1} e_{ik}\right\} \quad (4)$$

where $t_i$ means a tag and the $e_{ik}$ symbolize its fv-pairs and

$$p\left(\bigcap_{k=0}^{n-1} e_{ik} \middle| C_i\right) \cdot p(C_i) = p\left((\bigcap_{k=0}^{n-1} e_{ik}) \cap C_i\right)$$
$$= p(C_i) \cdot p(e_{i0}|C_i) \cdot p(e_{i1}|C_i \cap e_{i0}) \cdot$$
$$\ldots \cdot p\left(e_{i,n-1} \middle| C_i \cap \bigcap_{k=0}^{n-2} e_{ik}\right) \quad (5)$$

where $C_i$ means the context of $t_i$ and contains the tags $t_{i-2}$ and $t_{i-1}$ follows

$$p(t_i|C_i) = p(e_{i0}|C_i) \cdot \prod_{k=1}^{n-1} p\left(e_{ik} \middle| C_i \cap \bigcap_{j=0}^{k-1} e_{ij}\right) \quad (6)$$

The latter formula[3] describes the relation between the contextual probability of a tag and the contextual probabilities of its fv-pairs.

The unification of morphological features inside a noun phrase is accomplished indirectly. In a given context of fv-pairs the correct fv-pair obtains the probability $p=1$ and therefore will not influence the probability of the tag to which it belongs (e.g. $p_1^*($ 0num:SG $|...) = 1$ in fig. 2). A wrong fv-pair would obtain $p=0$ and make the whole tag impossible.

[3]suggested by Mats Rooth, IMS, Univ.Stuttgart, Germany

## 3 TRAINING ALGORITHM

In the training process we are not interested in analysing and storing the contextual probabilities (state transition probabilities) of whole tags but of single fv-pairs. We note them in terms of *probabilistic feature relations* (PFR):

$$\text{PFR} : \langle e_i \mid C_i^{sub} ; p(e_i|C_i^{sub}) \rangle \quad (7)$$

which later, in the tagging process, will be combined in order to obtain the contextual tag probabilities.

The term $e_i$ in formula (7) is a fv-pair. $C_i^{sub}$ is a *reduced* context which contains only a subset of the fv-pairs of a really appearing context $C_i$ (fig. 1). $C_i^{sub}$ is obtained from $C_i$ by eliminating all fv-pairs which do not influence the relative frequency of $e_i$, according to the condition:

$$p(e_i|C_i^{sub}) / p(e_i|C_i) \in [1-\epsilon, 1+\epsilon] \quad (8)$$

The considered fv-pair has nearly[4] the same probability in the complete and in the reduced contexts, i.e. $C_i$ does not supply more information about the probability of $e_i$ than $C_i^{sub}$ does.

(1a)

$t_{i-2}$ $\qquad$ $t_{i-1}$ $\qquad$ $t_i$

$$\begin{bmatrix} 2pos:DET \\ 2typ:DEF \\ 2gen:FEM \\ 2num:SG \end{bmatrix} \begin{bmatrix} 1pos:NOUN \\ 1gen:FEM \\ 1num:SG \end{bmatrix} \begin{bmatrix} 0pos:ADJ \\ \underline{0gen:FEM} \\ 0num:SG \end{bmatrix}$$

(1b)

$$\begin{bmatrix} \phantom{XXX} \end{bmatrix} \begin{bmatrix} 1gen:FEM \end{bmatrix} \begin{bmatrix} 0pos:ADJ \\ \underline{0gen:FEM} \end{bmatrix}$$

Figure 1: (a) Complete context $C_i$ and (b) reduced context $C_i^{sub}$ of the feature-value-pair $e_i = $ *0gen:FEM*

In the example (fig. 1a) we consider the fv-pair *0gen:FEM*. Within the given training corpus, its probability in the complete context $C_i$, i.e. in the context of all the other fv-pairs of figure 1a, is $p_0^*=44/44=1$ (cf. $p_0^*$ in fig. 2).

The presence of *1num:SG* in tag $t_{i-1}$ does not influence the probability of *0gen:FEM* in tag $t_i$. Therefore *1num:SG* can be eliminated. Only fv-pairs which really have an influence remain in the context. The reduced context $C_i^{sub}$ with less fv-pairs, which we obtain this way, is more general (fig. 1b).

In the given training corpus, the probability of *0gen:FEM* in the context $C_i^{sub}$ is $p_0=170/174=0.997$ (cf. $p_0$ in PFR$_0$ in fig. 2), which is near to $p_0^*=1$. The reduced context $C_i^{sub}$ is used to form a PFR which will be stored.

[4]A small change in the probability caused by the elimination of fv-pairs from the context is admitted if it does not exceed a defined small percentage $\epsilon$. (We used $\epsilon = 3\%$.)

We see in the use of reduced contexts instead of complete ones two advantages:

(1) A great number of complete contexts containing many fv-pairs can lead after elimination of irrelevant fv-pairs to the same PFR, which makes the number of all possible PFRs much smaller than the number of all possible trigrams (cf. sec. 2).

(2) The probability of a fv-pair can be estimated more exactly in a reduced context than in a complete one because of the higher frequencies in the first case.

**The Generation of PFRs**

In the training process we first extract from a training corpus a set of trigrams where the tags are split up into their fv-pairs. From these trigrams a set of PFRs is generated separately for every fv-pair $e_i$. We examined four different methods for this procedure:

**Method 1-3:** For every trigram we generate all possible subsets of its fv-pairs. Many trigrams, e.g. if they differ in only one fv-pair, have most of their subsets of fv-pairs in common. Both the complete trigrams and the subsets, constitute together the set of contexts and subcontexts ($C_i$ and $C_i^{sub}$) wherein a fv-pair could appear. To generate PFRs for a given fv-pair, we preselect and mark those (sub-)contexts which are supposed to have an influence on the contextual probability of the fv-pair. A (sub-)context will not be preselected if its frequency is smaller than a defined threshold. We use different ways for the preselection:

*Method 1:* A (sub-)context will be preselected if the considered fv-pair itself or an fv-pair belonging to the same feature type ever appears in this (sub-)context. E.g., if *gen:MAS* appears in a certain (sub-)context then this (sub-)context will be preselected for *gen:FEM* too. Furthermore, it is possible to impose special conditions on the preselection, e.g. that a (sub-)context can only be preselected if it contains a POS feature in tag $t_i$ and $t_{i-1}$ (cf. fig. 1a: *0pos* and *1pos*).

*Method 2:* In order to preselect (sub-)contexts for an fv-pair, we generate a decision tree[5] (Quinlan, 1983) where the feature of the fv-pair, e.g. *gen*, *num* etc., serves to classify all existing (sub-)contexts. E.g., *num* produces three classes of contexts: those containing the fv-pair *0num:SG*, those with *0num:PL* and those without a *0num* feature. We assign to the tree nodes other features than this upon which the classification is based. The root node is labeled with the feature from which we expect most information about the probability of the currently considered feature. The values of the root node feature are assigned to the branches starting at the root node. We continue the branching until there remain no features with an expected information gain and a frequency higher than defined thresholds. To every leaf of the tree corresponds a (sub-)context which will be marked and thus preselected for further analysis.

---

[5]suggested by Helmut Schmid, IMS, Univ. Stuttgart, Germany. For reasons of space we explain only how we employ decision trees for our purposes. For details about the automatic generation of such trees see Quinlan (1983).

*Method 3:* For each fv-pair concerning POS we preselect every (sub-)context containing only POS features in tag $t_{i-2}$ and $t_{i-1}$ (classical POS trigram), e.g. *2pos:PREP 1pos:DET* for *0pos:NOUN*. For the other fv-pairs we mark every (sub-)context containing any fv-pair of the same type in the previous tag $t_{i-1}$ and any POS features in tag $t_{i-1}$ and $t_i$, e.g. *1pos:DET 1gen:FEM 0pos:NOUN* for *0gen:FEM*.

With the methods 1-3, we next eliminate from every preselected (sub-)context all fv-pairs which in the above described sense do not influence the relative frequency of the currently considered fv-pair (eq. 8).

**Method 4:** From the set of trigrams extracted from a training corpus we generate separately for every fv-pair, a binary-branched decision tree which shall describe various contextual probabilities of this fv-pair. The tree is generated on a modified version of the ID3 algorithm (Quinlan, 1983) and is similar to the one described by Schmid (1994).

We start with a binary classification of all trigrams based on the considered fv-pair. E.g., a classification for *gen:FEM* will divide the set of trigrams in two subsets, one where the trigrams contain *0gen:FEM* in the tag $t_i$ and one where they do not.

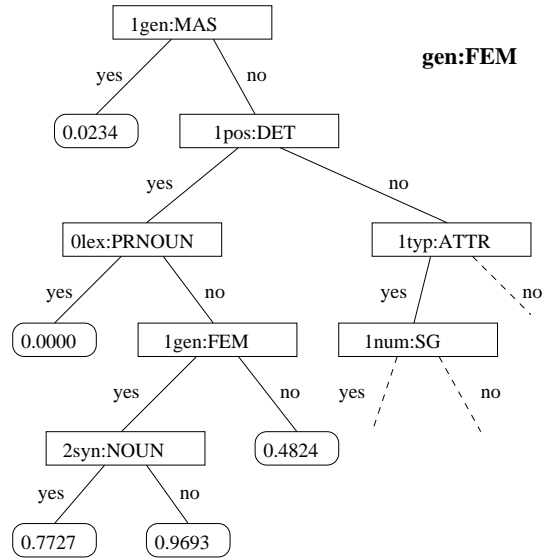

Figure 3: Decision tree for the fv-pair *0gen:FEM* (Every number is a probability of *0gen:FEM* in the context described by the path from the root node to the node labeled with the number.)

The tree is built up recursively (fig. 3). At each step, i.e. with the construction of each node, we test which one of the other fv-pairs delivers most information concerning the above-described classification. The current node will be labeled with this fv-pair. One of its two branches concerns the trigrams which con-

$$p(\ 0\text{gen:FEM}\ 0\text{num:SG}\ 0\text{pos:ADJ}\ |\ 1\text{gen:FEM}\ 1\text{num:SG}\ 1\text{pos:NOUN}$$
$$2\text{gen:FEM}\ 2\text{num:SG}\ 2\text{pos:DET}\ 2\text{typ:DEF}) = 44/298 = \underline{0.148}$$

$$p_0^*(\ 0\text{gen:FEM}\ |\ 0\text{num:SG}\ 0\text{pos:ADJ}\ 1\text{gen:FEM}\ 1\text{num:SG}\ 1\text{pos:NOUN}$$
$$2\text{gen:FEM}\ 2\text{num:SG}\ 2\text{pos:DET}\ 2\text{typ:DEF}) = 44/44 = 1.0$$
$$\textbf{PFR}_0 : \langle\ 0\text{gen:FEM}\ |\ 0\text{pos:ADJ}\ 1\text{gen:FEM}\ ;\ p_0 = 170/174 = 0.977\rangle$$

$$p_1^*(\ 0\text{num:SG}\ |\ 0\text{gen:FEM}\ 0\text{pos:ADJ}\ 1\text{gen:FEM}\ 1\text{num:SG}\ 1\text{pos:NOUN}$$
$$2\text{gen:FEM}\ 2\text{num:SG}\ 2\text{pos:DET}\ 2\text{typ:DEF}) = 44/44 = 1.0$$
$$\textbf{PFR}_1 : \langle\ 0\text{num:SG}\ |\ 0\text{pos:ADJ}\ 1\text{num:SG}\ 2\text{pos:DET}\ ;\ p_1 = 96/96 = 1.0\rangle$$

$$p_2^*(\ 0\text{pos:ADJ}\ |\ 1\text{gen:FEM}\ 1\text{num:SG}\ 1\text{pos:NOUN}$$
$$2\text{gen:FEM}\ 2\text{num:SG}\ 2\text{pos:DET}\ 2\text{typ:DEF}) = 44/298 = 0.148$$
$$\textbf{PFR}_2 : \langle\ 0\text{pos:ADJ}\ |\ 1\text{gen:FEM}\ 1\text{pos:NOUN}\ 2\text{pos:DET}\ ;\ p_2 = 69/465 = 0.148\rangle$$

$$\prod_{i=0}^{2} p_i = \underline{0.145}$$

The position index at the beginning of every feature-value-pair indicates the tag to which it belongs; e.g. *0gen:FEM* belongs to tag $t_i$ and *2num:SG* to $t_{i-2}$.

Figure 2: Decomposition and reconstruction of a contextual tag probability (state transition probability) using *probabilistic feature relations* (PFR)

tain the fv-pair, the other branch concerns the trigrams which do not contain it. The recursive expansion of the tree stops if either the information gained by consulting further fv-pairs or the frequencies upon which the calculus is based are smaller than defined thresholds.

## 4 TAGGING ALGORITHM

Starting point for the implementation of a feature structure tagger was a second-order-HMM tagger (trigrams) based on a modified version of the Viterbi algorithm (Viterbi, 1967; Church, 1988) which we had earlier implemented in C (Kempe ,1994). There we replaced the function which estimated the contextual probability of a tag (state transition probability) by dividing a trigram frequency by a bigram frequency (eq. 3) with a function which accomplished this calculus either using PFRs in the above-described way (eq.s 6, 7) or by consulting a decision tree (fig. 3).

To estimate the contextual probability of a tag we have to know the contextual probabilities of its fv-pairs in order to multiply them (eq. 6).

Using PFRs generated by **method 1 or 2**, when e.g looking for the probability $p_2^*(0\text{pos:ADJ}\ |...)$ from figure 2, we may find in the list of PFRs, instead of a PFR which would directly correspond (but is not stored), the two PFRs

$\langle 0\text{pos:ADJ}\ |\ 1\text{gen:FEM}\ 1\text{pos:NOUN}\ 2\text{pos:DET};$
$$p_1 = 0.148\rangle$$
$\langle 0\text{pos:ADJ}\ |\ 0\text{num:SG}\ 1\text{num:SG}\ 1\text{syn:NOUN}\ 2\text{syn:DET};$
$$p_2 = 0.414\rangle$$

Both of them contain subsets of the fv-pairs of the required complete context and could therefore both be applied. In such case we need to know how to combine $p_1$ and $p_2$ in order to get $p$ ($=p_2^*$ in fig. 2).

As there exists no mathematical relation between these three probabilities, we simply average $p_1$ and $p_2$ to get $p$ because this gives as good tagging results as a number of other more complicated approaches which we examined.

PFRs generated by **method 3** do not create this problem. For every complete context only one PFR is stored.

When we use the set of decision trees generated by **method 4**, we obtain for every fv-pair in every possible context only one probability by going down on the relevant branches until a probability information is reached.

In opposition to the PFRs of the other methods, the decision trees also contain negative information about the context of an fv-pair, i.e. not only which fv-pairs have to be in the context but also which ones must be absent.

## 5 TAGGING RESULTS

In the training and tagging process we experimented with different values for parameters like: minimal admitted frequency for preselection, admitted percentual difference $\epsilon$ between probabilities considered to be equal, etc. (cf. sec. 3).

The feature structure tagger was trained on the French 10,000 words corpus already mentioned in table 1, with the four different training methods (sec. 3). When tagging a 6,000 words corpus[6] with an average ambiguity of 2.63 tags per word (after the dictionary

---
[6]No overlap between training and test corpora.

look-up) we obtained in the best case an accuracy of 88.89 % (table 2).

| tag-ger | training corpus | | tag set | | HMM order | tagging accuracy |
|---|---|---|---|---|---|---|
| | number of words | language | tags | fv-prs. | | |
| tT | 2,000,000 | English | 47 | — | 1 | 94.93 % |
| tT | 2,000,000 | English | 47 | — | 2 | 96.16 % |
| tT | 10,000 | French | 386 | 57 | 1 | 56.39 % |
| tT | 10,000 | French | 386 | 57 | 2 | **83.23 %** |
| lpT | 10,000 | French | 386 | 57 | 2 | **83.81 %** |
| fsT1 | 10,000 | French | 386 | 57 | — | **88.53 %** |
| fsT2 | 10,000 | French | 386 | 57 | — | **88.89 %** |
| fsT3 | 10,000 | French | 386 | 57 | — | **88.44 %** |
| fsT4 | 10,000 | French | 386 | 57 | — | **88.14 %** |

$tT \rightarrow$ "traditional" HMM-tagger,
$lpT \rightarrow$ "Tagger" considering only lexical probabilities,
$fsT1..4 \rightarrow$ feature structure tagger
trained with method 1..4,
HMM order $1 \rightarrow$ bigrams, $2 \rightarrow$ trigrams

Table 2: Comparison of the tagging accuracy with different taggers, corpora, tag sets and HMM orders

Comparatively, we used a "traditional" HMM-tagger (cf. sec. 4) on the same training and test corpora and got an accuracy of 83.23 % [7], i.e. the error rate was about 50 % higher than with the feature structure tagger (table 2).

When we used a tool which always selects the lexically most probable tag without considering the context we obtained an accuracy of 83.81 %, which is even better than with the "traditional" HMM-tagger.

Provided with enough training data and working on a small tag set, our "traditional" tagger got an accuracy of 96.16 % (Kempe ,1994), which is usual in this case (Cutting et al.,1992). The English test corpus we used here had an average ambiguity of 2.61 tags per word which is amazingly similar to the ambiguity of the French corpus.

The feature structure tagger is clearly better when the available training corpus is small and the tag set large but the tags are decomposable into few fv-pairs.

# 6 FURTHER RESEARCH

We intend to search for other similar models while keeping in mind the basic idea described above: Splitting up a tag into fv-pairs and deducing its contextual probability from the contextual probabilities of its fv-pairs.

Furthermore, it may be preferable to split up the tags only when the frequencies are too small[8].

---

[7] For a similar experiment for German (20,000 words training corpus, 689 tags, trigrams) an accuracy of 72.5 % has been reported (Wothke et al., 1993, *p. 21*).

[8] suggested by Ted Briscoe, Rank Xerox Research Centre, Grenoble, France